\documentclass[a4paper, 10pt, conference]{ieeeconf}      

\IEEEoverridecommandlockouts                              

\usepackage{graphicx} 
\usepackage{amsmath}
\usepackage{amssymb}

\usepackage{bm}
\usepackage{bbm}
\usepackage{dsfont}
\usepackage{siunitx}
\DeclareMathOperator*{\argmax}{arg\,max}
\usepackage{tikz}
\usepackage{pgfplots}					
\usepackage{pgfplotstable}
\usepgfplotslibrary{statistics}		
\pgfplotsset{width=7cm,compat=newest, every tick label/.append style={font=\tiny}}	
\usepackage{filecontents}
\usepackage{todonotes}
\usepackage{pdfpages}
\usepackage{pgf}
\usepackage{cite} 	
\usetikzlibrary{external}
\title{\LARGE \bf
Radar-based Road User Classification and Novelty Detection with Recurrent Neural Network Ensembles
}

\author{Nicolas Scheiner$^{1}$, Nils Appenrodt$^{1}$, J\"urgen Dickmann$^{1}$, and Bernhard Sick$^{2}$
\thanks{$^{1}$Daimler AG, Wilhelm-Runge-Str. 11, 89081 Ulm, Germany
        {\tt\small nicolas.scheiner@daimler.com}}
\thanks{$^{2}$Intelligent Embedded Systems, University of Kassel, Wilhelmsh\"oher Allee 73, 34121 Kassel, Germany
        {\tt\small bsick@uni-kassel.de}}%
}

\begin{document}

\maketitle
\thispagestyle{plain}
\pagestyle{plain}

\begin{abstract}
Radar-based road user classification is an important yet still challenging task towards autonomous driving applications.
The resolution of conventional automotive radar sensors results in a sparse data representation which is tough to recover by subsequent signal processing.
In this article, classifier ensembles originating from a one-vs-one binarization paradigm are enriched by one-vs-all correction classifiers.
They are utilized to efficiently classify individual traffic participants and also identify hidden object classes which have not been presented to the classifiers during training.
For each classifier of the ensemble an individual feature set is determined from a total set of 98 features.
Thereby, the overall classification performance can be improved when compared to previous methods and, additionally, novel classes can be identified much more accurately.
Furthermore, the proposed structure allows to give new insights in the importance of features for the recognition of individual classes which is crucial for the development of new algorithms and sensor requirements.
\end{abstract}

\section{Introduction} \label{seq:intro}
Radar sensing is an integral part of many perception concepts for autonomously driving vehicles.
This is justified by a radar's ability to directly obtain a precise radial (Doppler) velocity from all observed objects within a single measurement.
It is currently the only automotive sensor to deliver this information in a single shot fashion and, therefore, is indispensable for adaptive cruise control systems \cite{winner2016}.
Moreover, state-of-the-art automotive radar sensors typically operate at a frequency range of $76-$\SI{81}{GHz}.
This makes them more robust to adverse weather conditions such as fog, snow, or heavy rain.
The drawback of radar is its low angular resolution when compared to other sensors.
This leads to sparse data representations, especially for remote objects.

\begin{figure}[t!]
	\centering	
	\includegraphics[width=1.\columnwidth]{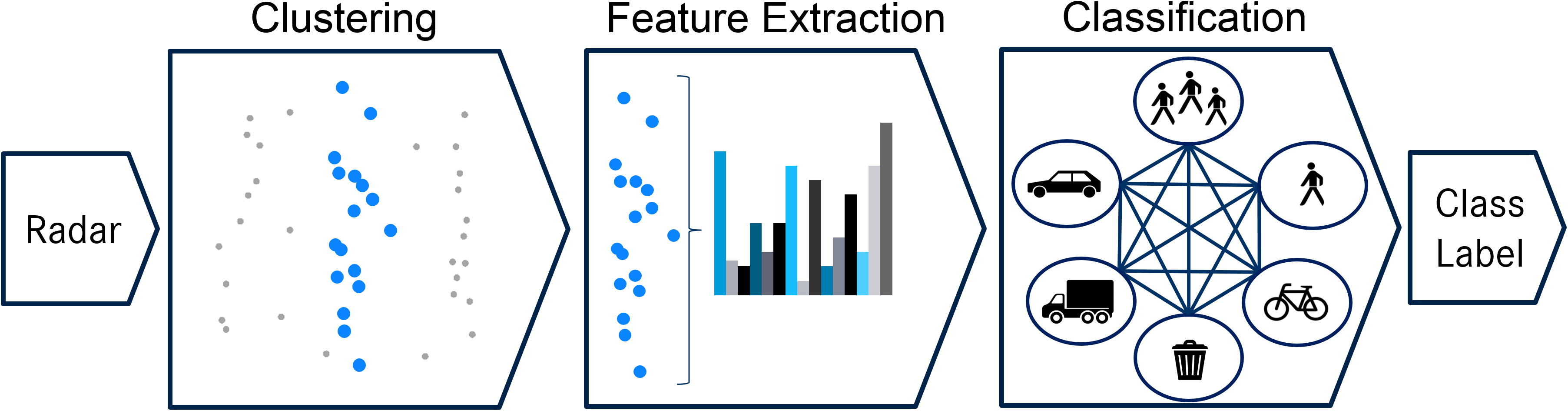}
	\caption{Conventional radar data classification process: data is first grouped and features calculated based on the clusters. In this case, a decomposed and combined one-vs-one (indicated by straight lines connecting classes) and one-vs-all (circles around classes) multiclass classifier is depicted.}	
	\label{fig:abs_clf_proc_old}
\end{figure}

Due to the high complexity of identifying object instances in these kinds of data, most of the research utilizes radar for gaining a basic first understanding about the scene at hand.
Hence, only few classes are separated from each other, e.g., \cite{Heuel2011, bartsch12, Sorowka2015}.
Excellent environmental perception requires understanding more complex scenes.
Thus, in \cite{Schumann2017} and \cite{Scheiner2018}, five classes of dynamic road users are recognized and separated from a sixth class comprising measurement artifacts and other undesired data points.
This article is based on the work presented in \cite{Scheiner2018}.
By applying different kinds of multiclass binarization techniques, the classification performance is improved, there.
Starting from a set of 50 features, a backward elimination routine is applied.
The resulting optimized set of 36 features leads to a major performance boost.
In accordance with conventional multiclass binarization strategies, the feature set is fixed for all individual classifiers composing the model (cf. Fig \ref{fig:abs_clf_proc_old}).
However, it has to be expected that a feature set optimized for the combined results on all classifiers of the overall model is inferior to the same ensemble in which each classifier is presented its own optimized feature set.
Finding an optimal feature set is an NP-hard problem, hence, even for one single feature set, estimating an optimal subset becomes infeasible with an increasing number of features \cite{Amaldi1998}.
In order to limit the computational effort, this article proposes a combined approach of a wrapper method (backward elimination) which is enriched by a heuristic originating from two filter methods (Joint Mutual Information and the Relief-based MultiSURF algorithm).
This allows to find a well performing feature set for each individual classifier in a reasonable time.

\begin{figure}[t!]
	\centering
	\includegraphics[width=1.\columnwidth]{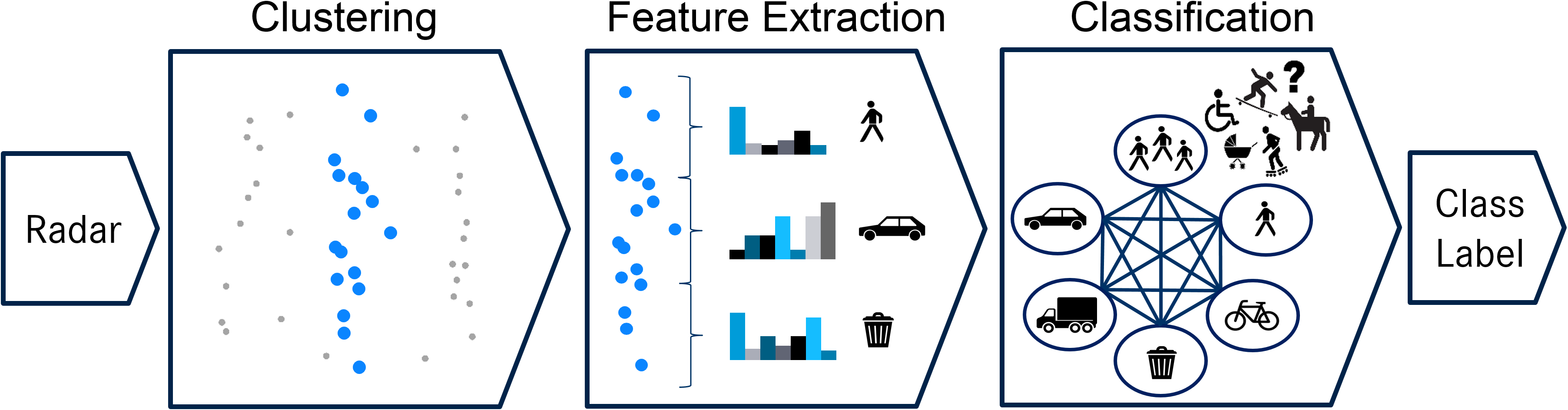}
	\caption{Proposed processing chain: each classifier in the ensemble is presented with its own specialized feature set. The results aggregation scheme accounts for previously unseen patterns.}	
	\label{fig:abs_clf_proc_new}
\end{figure}

An additional issue, that is often neglected in publications about automotive radar classification, is the handling of previously unseen classes.
This problem is critical for the path planning of the vehicle, i.e., estimating where the car can drive within adequate safety margins from the expected positions of other traffic participants and surroundings.
Therefore, it might be of minor concern to discriminate between, e.g., buses and  trucks.
However, the dynamics of a train is less similar to those examples, and a skateboarder moves differently than a regular pedestrian, although they look similar when observed by a radar sensor.
Classifiers are often forced to make a decision about the class membership of an object.
In the worst case scenario they might even decide that a novel data pattern is most likely a measurement artifact and can, therefore, be ignored.
These few examples illustrate the importance of giving the path planning stages in a vehicle accurate information about other road users.
Indeed, this issue is also present in other areas such as data acquisition:
As the number of available vehicles with advanced sensor setups increases, the data collection process for automotive data sets becomes easier.
However, methods for selecting interesting data require measures for the criticality or -- in this case -- novelty of the data.
To this end, this article describes how the result aggregation scheme for the binarized classifiers can be altered to yield and additional \emph{hidden} class.
Final results show clear benefits of the proposed methods.
The overall classification performance can be further increased, despite the more challenging data set.
The utilized long short-term memory (LSTM) cell classifier provides a lightweight network structure which has several advantages:
First of all, the small amount of parameters result in low computational efforts during model training and inference.
The short training time is due to the small amount of model parameters which have to be adjusted.
This leads to the second advantage which is the lower amount of training data required for the learning algorithm to converge without overfitting.
Third and last, given good clustering results in the preprocessing stages, the accuracy of the LSTM approach is exceedingly high.
A recent trend in machine learning is the incorporation of preprocessing and classification steps in a single convolutional neural network (CNN).
This has also been done for automotive radar classification, e.g., \cite{Schumann2018} or \cite{Gaisser2017}.
Due to the aforementioned reasons, the LSTM approach is preferred, here.

The article is organized as follows: In Section \ref{sec:data} the data set and essential preprocessing steps are described.
The first part of Section \ref{sec:methods} explains how suitable feature sets are selected.
The second part deals with the proposed aggregation scheme for novelty detection with binary classifiers.
Section \ref{sec:results} presents the results and Section \ref{sec:conclusion} concludes the topic and gives prospects for future work.

\section{Data Set And Preprocessing}\label{sec:data}
The data used in this article originates from a set of real-world automotive radar data as previously used in \cite{Schumann2018}.
It contains more than 3 million data points on roughly 3800 instances of moving road users.
The class membership of road users is distributed as depicted in Tab. \ref{tab:data}.
While most classes are self-explanatory, the label \emph{pedestrian group} is attributed to multiple pedestrians which cannot be clearly separated in the data.
Moreover, the \emph{garbage} class consists of wrongly detected and clustered measurement artifacts.
The hidden \emph{other} class is made up from several road users which do not strictly fit into any of the aforementioned groups.
\begin{table}[tb]
\renewcommand{\arraystretch}{1.3}
	\caption{Object Instance (Upper Row) And Sample (Lower Row) Distribution In Data Set.}
	\label{tab:data}
	\centering
	\begin{tabular}{ccccccc}
		\hline
 		Pedestrian & Group & Bike & Car & Truck & Garbage & Other \\
		\hline\hline
		1215 & 1063 & 90 & 1310 & 154 & 38176 & 22 \\
		\hline
		30623 & 46053 & 5541 & 33920 & 6648 & 69597 & 790 \\
		\hline
	\end{tabular}
\end{table}

\begin{figure}[t!]
	\centering
	\includegraphics[width=1.\columnwidth]{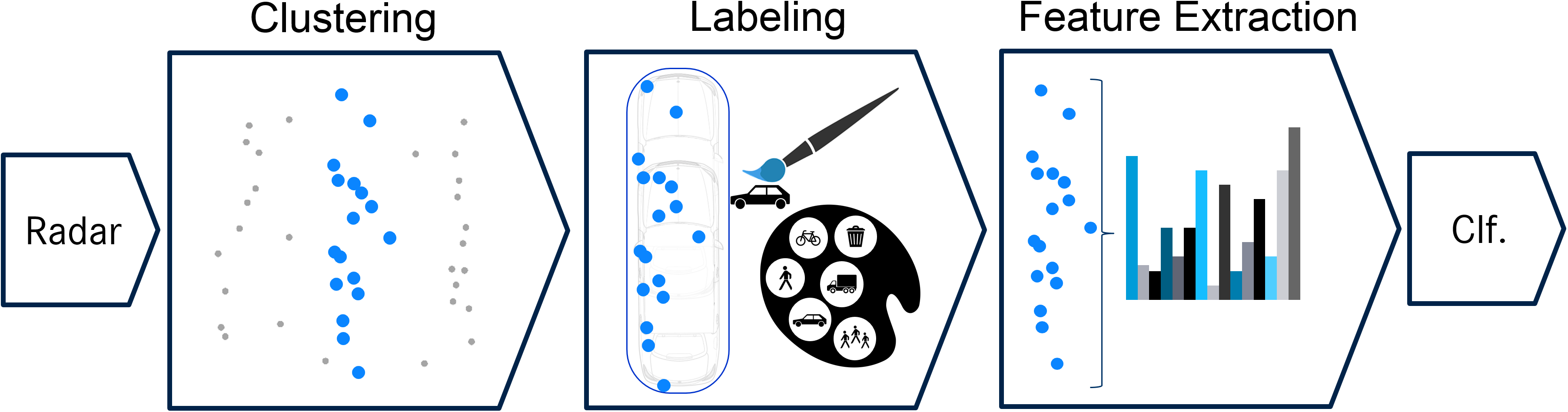}
	\caption{Data preprocessing overview: data is first transformed to common coordinate system, then it is clustered and labeled before extracted features are presented to the classification model.}	
	\label{fig:clf_chain}
\end{figure}

All data was acquired with four radar sensors distributed over the front half of a test vehicle.
Due to data sparsity, overlapping regions in the sensors' field of view do not get any special treatment in the accumulation process.
The sensors operate at \SI{77}{\giga\hertz} carrier frequency.
Their beams span from \SI{-45}{\degree} to \SI{+45}{\degree} and up to a range of \SI{100}{\meter} away from the sensor.
The radars deliver points on a detection level which are already resolved in range, angle, and radial Doppler velocity.
They are prefiltered by an internal constant false alarm rate filter and are clustered in space, time, and Doppler using a customized DBSCAN \cite{Ester1996} algorithm.
In an annotation process, each cluster is assigned a class label. Cluster refinement ensures that all detections associated with the object are being considered.
Uncorrected and intentionally deteriorated versions of the clustered data are used for data augmentation purposes during training.
For feature extraction, all labeled cluster sequences are first sampled in time using a window of $\SI{150}{ms}$.
Then, features are extracted from each of the cluster samples.
Up to eight feature vectors are being concatenated in a sliding window fashion in order to form input sequences to the neural network used for classification.
The preprocessing stage is summarized in Fig. \ref{fig:clf_chain}.
More details on all involved steps are given in \cite{Scheiner2018}.

\section{Methods}\label{sec:methods}
The underlying classifier units used for this article are long short-term memory (LSTM) cells.
LSTMs are a special kind of recurrent neural network which introduce gating functions in order to avoid the vanishing gradient problem during training \cite{hochreiter97}.
They perform well on time series of all kinds and provide a straightforward way to utilize the time information in the data.
A fixed configuration of 80 LSTM cells followed by a softmax layer is used for all classifiers of the ensemble which will be described in the following subsection.
Experiments with larger LSTM layers of 120 and 200 cells did not indicate any relevant improvements.

\subsection{Ensemble Creation With Classifier Specific Feature Sets}
As previously shown in \cite{Scheiner2018}, multiclass binarization is a well-equipped technique for improving the classification performance on moving road users.
Best results were achieved by using a combined one-vs-one (OVO) and one-vs-all (OVA) approach.
Class membership is estimated by summing all pairwise class posteriors probabilities $p_{ij}$ from corresponding OVO classifiers.
Thereby, each OVO classifier is weighted by the sum of corresponding OVA classifier outputs $p_{i}$.
Subscripts $i$ and $j$ denote the corresponding class ids for which the classifier was trained.
During testing, this limits the influence of OVO classifiers which were not trained on the same class as the regarded sample, i.e., the OVA classifiers act as \emph{correction classifiers} \cite{Moreira1998}.
The final class decision for a feature vector $\bm{x}$ can then be calculated as:
\begin{align}
	\texttt{id}(\bm{x}) = \argmax_{i \in \{1,...,K\}} \text{\hspace{1mm}} \sum_{j=1, j\neq i}^K p_{ij}(\bm{x}) \cdot (p_{i}(\bm{x}) + p_{j}(\bm{x}))\texttt{.}
	\label{eq:opc}
\end{align}
$K$ is the amount of classes in the training set.
This combined approach of OVO and OVA yields a total of $K(K+1)/2$ classifiers.
Traditionally each classifier in the decomposed model uses the same feature set as input.
It is, however, most likely that not every feature is equally important for each classifier.
Moreover, early experiments with new features revealed that feature selection algorithms did not perform very well on larger feature sets.
The methods were either reluctant to reduce the complexity and drop features at all, or did not result in beneficial configurations in reasonable time.
Thus, this article proposes a method for estimating individual well performing feature sets for each binary classifier.

Before the selection algorithm is described, an overview over the features under consideration is necessary.
In \cite{Scheiner2018} a set of 50 features was described.
Each feature is calculated from all detections in a cluster during a certain time frame.
The features can be roughly divided into three groups:
statistical derivations of the four base units (range, angle, amplitude, and Doppler), geometric features describing the spatial distribution of detections in a cluster sample, and features concerning the micro-Doppler characteristics, i.e., the distribution of Doppler values on the observed object.
For this article, the set of possible feature candidates is further increased.
A variety of features was examined in \cite{Wagner2018}, however, due to redundancies only a couple of features can be transferred to this work.
For instance, the mean distance between all pairs of detections is adopted as a new measure of compactness.
Furthermore, the previous DBSCAN clustering stage can also deliver an amount of core samples besides grouping points.
The ratio of core points to the total number of detections serves as additional feature.
Inspired by the findings in \cite{volca2017}, the resemblance of the detection distribution to an ellipse shape is estimated.
Therefore, the ratio of points close to the estimated outlines of the object to the total amount of detections is calculated.
Moreover, the eigenvalues of the covariance matrix of all the x/y coordinates, Doppler velocities, and amplitudes are used, along with some additional nonlinear transformations of important base values.
A complete list of all 98 considered features can be found in Tab. \ref{tab:feats} in the appendix.

In order to determine optimized feature sets for all classifiers in the ensemble, feature selection techniques have to be applied.
The two most popular variants of feature selection techniques are \emph{filter} and \emph{wrapper} methods.
Filter methods rank features by calculating statistical relevance measures, e.g., the correlation between individual features and class labels.
This can be done independently from any classification algorithm and is computationally efficient.
On the downside, many filter techniques do not consider interdependency between features and are, therefore, suboptimal.
In comparison, wrapper methods use an actual classifier to test different subsets of features.
Most important variants of wrapper methods include \emph{backward elimination} and \emph{forward selection}.
Both methods greedily eliminate or add the one best fitting feature at a time without going back until some stop criterion is reached.
While those methods can usually find better subsets than filter methods, their computational costs are very high.
A more detailed overview over feature selection methods can be found, e.g., in \cite{BENNASAR20158520}.

The high amount of features and classifiers does not allow to repeatedly perform a full backward elimination sweep.
Hence, a combined approach of filter and wrapper methods is used in order to get a suitable subset in reasonable time.
Essentially, instead of testing all features in a backward elimination run for the least well performing one, only one feature is tested at a time.
If test results reveal a benefit of dropping the regarded feature, it is removed from the feature set.
Otherwise, the next feature is evaluated.
In order to determine the order in which to examine features, a feature ranking originating from the combination of two filter methods is used as a heuristic.

The \emph{Joint Mutual Information} (JMI) criterion is the first utilized filter.
JMI is a method that iteratively adds the feature maximizing the combined amount of information that reduces the uncertainty of the underlying class \cite{yang1999}.
It is a very popular variant due to its low complexity despite high stability.
As a second heuristic the \emph{MultiSURF} algorithm is used to create a ranking.
MultiSURF originates from the family of \emph{Relief-based algorithms}.
In contrast to the JMI criterion, Relief-based algorithms aim to build feature weights -- which can be interpreted as ranking -- based on proximity measures between different samples in the data set \cite{URBANOWICZ2018189}.
In doing so, Relief-based algorithms better account for interdependence between multiple features.

\begin{figure}[htb]
	\centering
	\includegraphics[width=0.9\columnwidth]{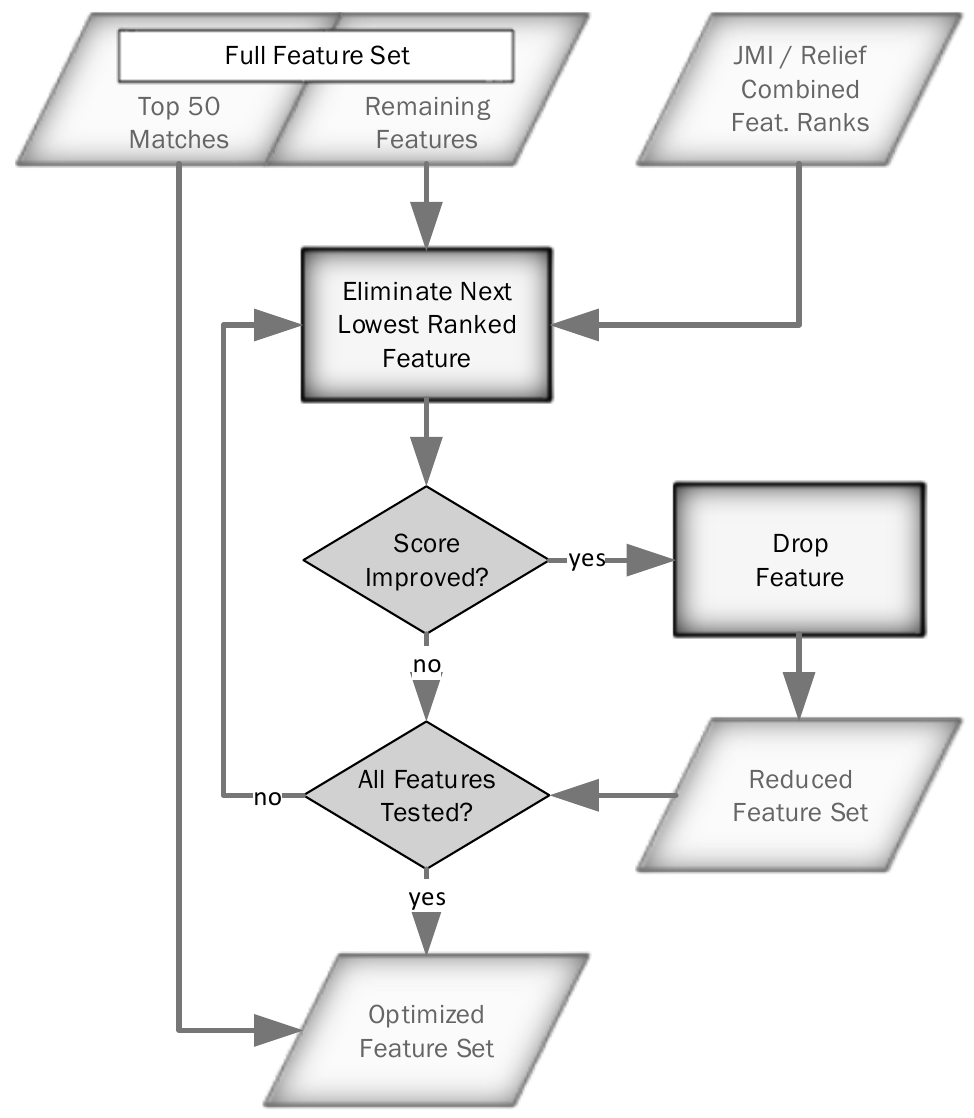}
	\caption{Simplified workflow of feature selection algorithm. A backward elimination scheme is enriched by a heuristic to successively eliminate bad-performing features. If no more features can be eliminated without decreasing the classification performance, the so-found subset is added to fixed set determined by previous feature ranking.}
	\label{fig:feat_sel}
\end{figure}

In order to profit from both feature rating approaches, the individual rankings of both methods are averaged.
The shared features in both rankings' top 50 are directly added to the final feature set to further decrease the computational costs.
The remainder of the combined ordering is used as heuristic for a directed backward elimination process as depicted in Fig. \ref{fig:feat_sel}.
This process is repeated for every classifier in the ensemble.

\subsection{Hidden Class Detection}
Another aim of this article is giving the classification ensemble the ability to detect hidden classes that have not been previously presented to the classifier.
To this end, three different approaches are being examined:
\paragraph{OVA thresholding} The OVA classifiers of the ensemble are already being used for weighting corresponding OVO classifiers based on their estimated relevancy.
This way, a \emph{relevancy threshold} can be defined which determines if any of the OVO classifiers is likely to be able to make a competent decision about the class membership of the sample.
Otherwise, the hidden class label is chosen if:
\begin{align}
	p_{i}(\bm{x}) < \text{thr} \text{\hspace{5mm}} \forall i \in \{1,...,K\}\texttt{.}
	\label{eq:ovathr}
\end{align}
\paragraph{Voting} In order to incorporate the whole ensemble in the decision process, a voting scheme was proposed in \cite{Chang2011}.
Hidden classes are assumed if no class label gets a minimum number of votes, i.e.:
\begin{align}
	\mathds{1}[p_{i}(\bm{x})>0.5] + \sum_{j\neq i} \mathds{1}[p_{ij}(\bm{x})>0.5] < \text{thr} \text{\hspace{5mm}} \forall i \texttt{.}
	\label{eq:ovathr}
\end{align}
$\mathds{1}[\mathord{\cdot}]$ denotes the indicator function.
\paragraph{OVO+OVA thresholding} Instead of finding the maximum value for the combined OVO and OVA approach in Eq. \ref{eq:opc}, the K class scores can be normalized to form pseudo probabilities.
These probabilities are then compared to a threshold value similar to the other two methods:
\begin{align}
	\nonumber &c^{-1} \cdot \sum_{j\neq i} p_{ij}(\bm{x}) \cdot (p_{i}(\bm{x}) + p_{j}(\bm{x})) < \text{thr} \text{\hspace{5mm}} \forall i \\
	&\text{\hspace{1mm} with \hspace{5mm}} c := \sum_{i} \sum_{j\neq i} p_{ij}(\bm{x}) \cdot (p_{i}(\bm{x}) + p_{j}(\bm{x}))\texttt{.}
	\label{eq:opcthr}
\end{align}

If parameterized correctly, all three methods are able to identify hidden classes with a reasonable performance.
The used data set only contains very few examples of other road users.
It includes motorcyclists, scooters, wheelchair users, cable cars, and dogs.
The main challenge is, however, to determine a parameter setting which has low impact on the correct classification of the existing classes.
In the next section, all three techniques will be evaluated based on their detection performance of the hidden class and their impact on the regular six classes.

\section{Results}\label{sec:results}
The data set used in this work contains strong imbalances among individual class occurrences.
In order to preserve the influence of each individual class, all classification scores are reported as \emph{macro-averaged F1 scores} where not mentioned otherwise.
F1 is defined as the harmonic mean of precision (true positive / predicted positive) and recall (true positive / condition positive).
Macro-averaging obtains F1 scores for each individual class, then uses the mean value of all $K$ F1 scores ($F1_\text{macro}=\frac{1}{K}\sum_{i=1}^K F1_i$) as a final measure.

Suitable feature sets are determined from the averaged scores based on 5-fold cross-validation.
During the guided backward elimination process each feature is tested with five different combinations of train and test data.
The averaged results determine, whether a feature is kept or removed from the set.
The so estimated classifier-specific feature sets are maintained for further experiments.

As it has to be expected that classifying a hidden class besides the original class set is more complicated, the results for the six class problem with and without the hidden class are discussed separately.
Moreover, an interpretation of the chosen feature sets is given at the end of the section.

\subsection*{Classification Performance On Six Class Problem}
For the evaluation of the proposed feature selection strategy on the original six class classification problem, it is necessary to define some baseline experiments for comparative reasons.
Three possible candidates utilize a shared feature set in addition to the combined decomposed OVO/OVA ensemble.
The common feature sets are, first, the full set of 98 features.
Second, a feature reduction utilizing the proposed guided backward elimination routine on the whole ensemble, and third, the optimized feature set determined in \cite{Scheiner2018}.
Moreover, a multiclass classification approach with a single classifier is evaluated on the full feature set.
Due to the largely increased feature set, running a full backward elimination without feature ranking is not possible in reasonable time.
Stable results are ensured by a 10-fold cross-validation routine.
The presented scores are the averaged results as depicted in Tab. \ref{tab:scores}.

\begin{figure}[htb]
	\centering
	\includegraphics[width=0.9\columnwidth]{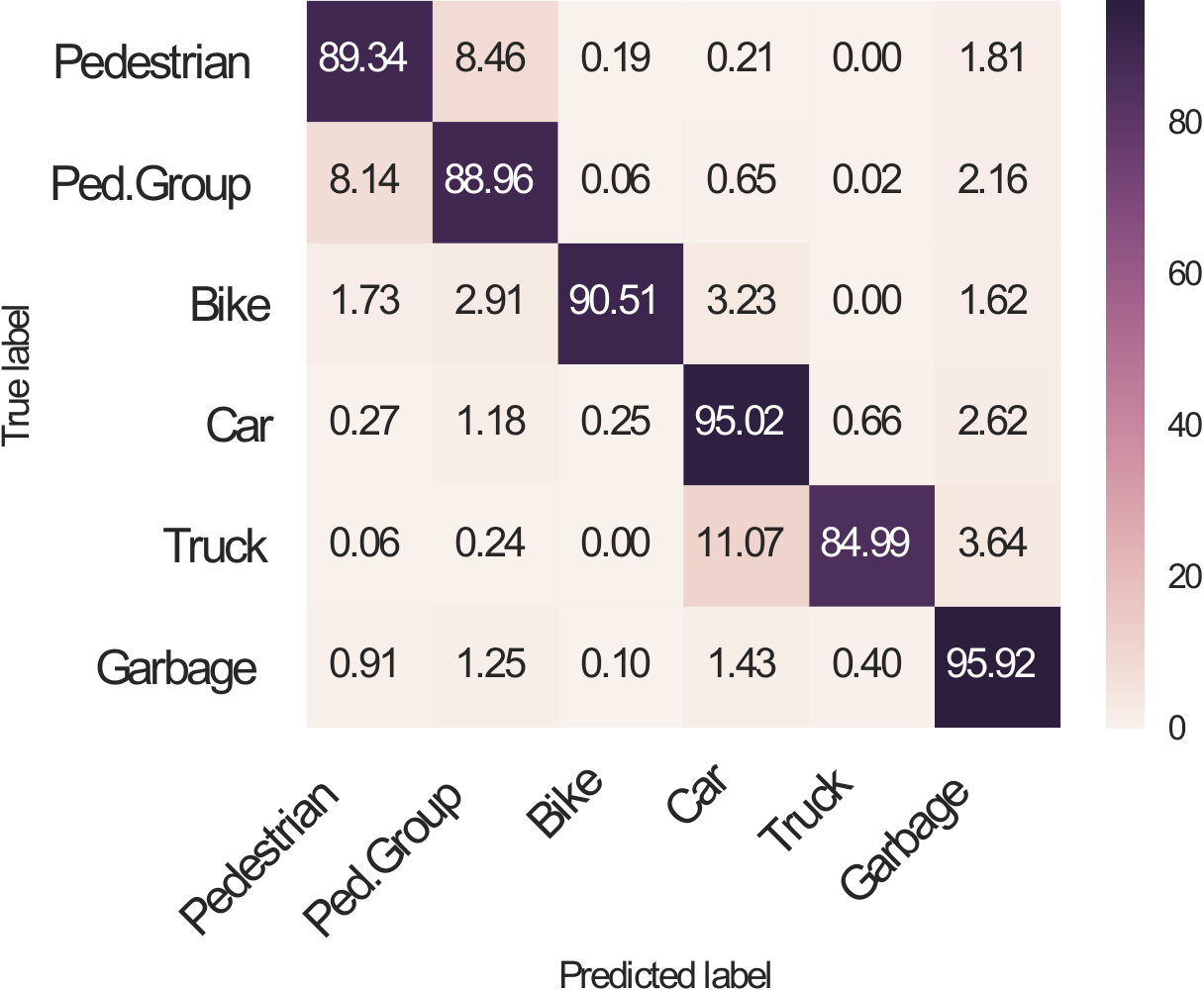} 
	\caption{Normalized confusion matrix. All values as percentages.}
	\label{fig:conf_6class}
\end{figure}

With a \textbf{final score of \SI{91.46}{\%}}, the proposed method demonstrates its effectiveness over other methods.
Considering the already high base score from \cite{Scheiner2018}, the improvement of \SI{0.33}{\%} is a remarkable increase.
Interestingly, the larger feature set did not lead to a performance improvement. 
Even in its reduced version after performing feature selection, the score could be only improved slightly, when the same feature set was shared between all classifiers.
This leads to the conclusion, that the complexity of optimizing the large feature set on the whole ensemble is simply too high for the utilized feature selection algorithm, hence, backing the classifier-specific selection approach.
As expected, the multiclass classification approach with a single classifier has the worst performance and will not be discussed any further.
In order to get a better understanding in the underlying classification subprocesses, the confusion matrix in Fig. \ref{fig:conf_6class} displays the distributions of predicted and true class memberships.
Most classification errors occur between the very similar \emph{pedestrian} and \emph{pedestrian group} classes.
Also, several cars are falsely classified as trucks.
Despite the problems with some similar classes, these issues are well compensated by the overall performance.

\begin{table}[htb]
\renewcommand{\arraystretch}{1.4}
	\caption{Classification Results On Six Class Problem.}
	\label{tab:scores}
	\centering
	\begin{tabular}{ll}
		\hline
 		Method & $\text{F1}_{\text{macro-averaged}}$ \\
 		\hline
 		\textbf{Proposed ensemble method} & \textbf{\SI{91.46}{\%}} \\
		Shared full feature set & \SI{91.08}{\%} \\
		Optimized shared features & \SI{91.17}{\%} \\
 		Shared feature set from \cite{Scheiner2018} & \SI{91.13}{\%}  \\
 		Full feature multiclass & \SI{90.64}{\%} \\
 		\hline
	\end{tabular}
\end{table}

\subsection*{Classification Results With Hidden Class}
\begin{figure}[tb]
\centering
\includegraphics{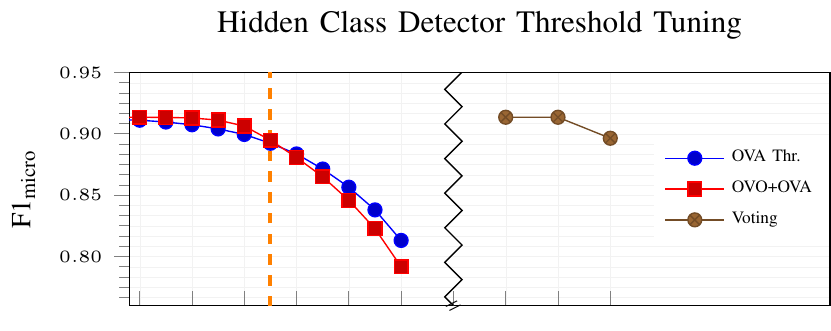}
\includegraphics{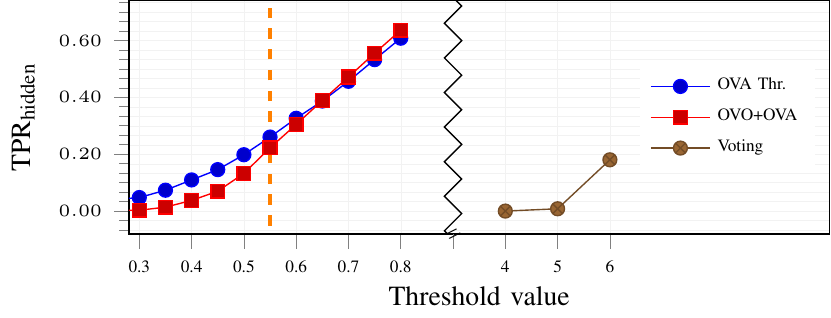}
\caption{Threshold estimation of the hidden class detector based on three different methods: classifier voting, OVA, and OVO+OVA thresholding. For each method, the true positive rate on hidden class detection and the micro-averaged F1 scores over all test samples is given at the bottom and top, respectively. The chosen threshold level is indicated as dashed orange line.}
\label{fig:hidden_class_tuning}
\end{figure}

\begin{figure}[htb]
	\centering
	\includegraphics[width=1.0\columnwidth]{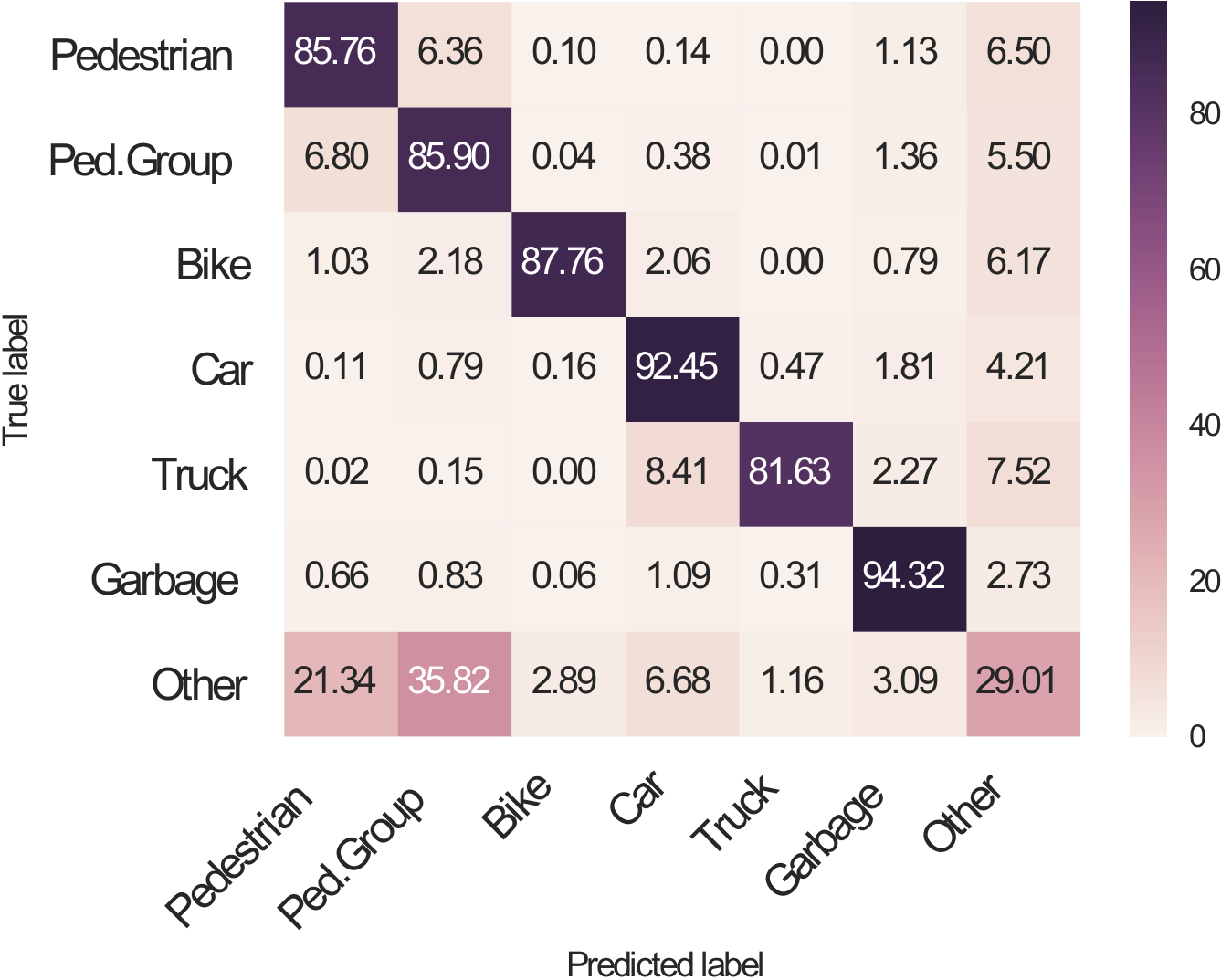} 
	\caption{Normalized confusion matrix. All values as percentages.}
	\label{fig:conf_7class}
\end{figure}

Before the classification scores of the model on the six class problem including a hidden class can be calculated, the detection thresholds for the hidden class need to be set first.
The parameterization is done empirically, i.e., experiments are evaluated several times with different threshold settings.
Therefore, the data samples forming the hidden class are split in two parts of roughly equal size.
Each road user instance may only occur in either of those splits to avoid biasing the results.
Experiments are conducted for all three variants discussed in Section \ref{sec:methods}.
The results of these test runs are shown in Fig. \ref{fig:hidden_class_tuning}.
The goal is to maximize the hidden class detection performance without tempering with the scores on the remaining six classes.
Therefore, Fig. \ref{fig:hidden_class_tuning} indicates the true positive rate (TPR) of the hidden class and the micro-averaged F1 score based on all test samples from either class.
Micro-averaging, i.e., averaging over all samples and not classes, is the preferred metric in this case as it allows to estimate the direct impact on all classes when detecting an additional class.
The experiments indicate that all three methods are able to detect an additional hidden class without extreme losses in classification scores.
Although, the voting strategy is limited in its parameterization (only 5 or 6 vote thresholds show any effect), it provides an easy way to reach up to \SI{18}{\%} TPR for the hidden class.
OVA and combined OVO+OVA thresholds can be freely adjusted which allows to settle for more optimized choice of parameters.
However, the combined OVO+OVA approach seems inferior to the pure OVA implementation fore almost all threshold settings.
Even at high thresholds, when the combined approach reaches higher TPRs on the hidden class, the total score on all classes decreases compared to OVA thresholding.
Additional experiments were conducted with the aim to combine two or even all three proposed detection schemes.
The results did, however, not indicate any relevant improvements.
For final reported scores, the \emph{OVA thresholding} method with a corresponding threshold of 0.55 is chosen.
Hereby, the macro-averaged F1 score of the trained six classes decreases to \textbf{\SI{90.83}{\%}}, which is still a very respectable value, given the fact, that the model can now decide for an additional class label that was not in the training data.
The hidden class is detected at a rate of \textbf{\SI{29.01}{\%}} which is not a lot but far better than not being able to detect it at all.
As shown in Tab. \ref{tab:scores_hidden}, the proposed method also outperforms other methods in terms of hidden class detection rate.
A confusion matrix for the results on the new model can be found in Fig. \ref{fig:conf_7class}.
Evidently, the major confusions are present for \emph{other class} samples being classified as pedestrians or pedestrian groups.
As a large part of this class is made up by wheelchair user -- sometimes accompanied by other pedestrians -- and scooter drivers, this confusion is comprehensible.
Obviously, in the near future hidden classes in automotive radar classification tasks will no longer include scooters or wheelchair users as those classes will have representative numbers in the training data by then.
However, it is not possible to model the whole world, hence, a hidden class detector will always be of relevance.

\begin{table}[htb]
\renewcommand{\arraystretch}{1.4}
	\caption{Classification Results With Hidden Class Detector.}
	\label{tab:scores_hidden}
	\centering
	\begin{tabular}{lll}
		\hline
 		Method & $\text{F1}^{\text{6class}}_{\text{macro-avg}}$ & $\text{TPR}_{\text{hidden}}$ \\
 		\hline
 		\textbf{Proposed ensemble method} & \textbf{\SI{90.83}{\%}} & \textbf{\SI{29.01}{\%}} \\
		Shared full feature set & \SI{90.62}{\%} & \SI{24.71}{\%} \\
		Optimized shared features & \SI{90.70}{\%} & \SI{24.76}{\%}  \\
 		Shared feature set from \cite{Scheiner2018} & \SI{90.47}{\%} & \SI{27.54}{\%} \\
 		\hline
	\end{tabular}
\end{table}

\subsection*{Evaluation Of Feature Subsets}
Besides improving the overall classification score, the proposed ensemble structure gives the opportunity to more closely evaluate the performance of different features.
More specifically, the impact of each feature on a smaller and distinct classification problem can be examined.
In order to get a better overview on the distribution of features on the individual classifiers, instead of discussing features individually, all features have been divided into six different groups.
The group definitions aim to summarize features that were designed with a common purpose, e.g., to better inform the classifier about a cluster amplitude distribution or shape.
Therefore, the first four groups include features that are supposed to reveal valuable characteristics based on the basic radar values: \emph{Doppler velocity}, \emph{amplitude}, \emph{range}, and \emph{angle}.
Another group describes the \emph{shape} or compactness of a cluster.
Lastly, the \emph{microdoppler} category summarizes the interaction between velocity and spatial information, i.e., how radial velocities are distributed over the cluster area.
As it is sometimes not perfectly clear, in which category to put a feature, the category assignment information for each feature is also given in the feature overview in the appendix.

\begin{figure}[b!]
\centering
\includegraphics{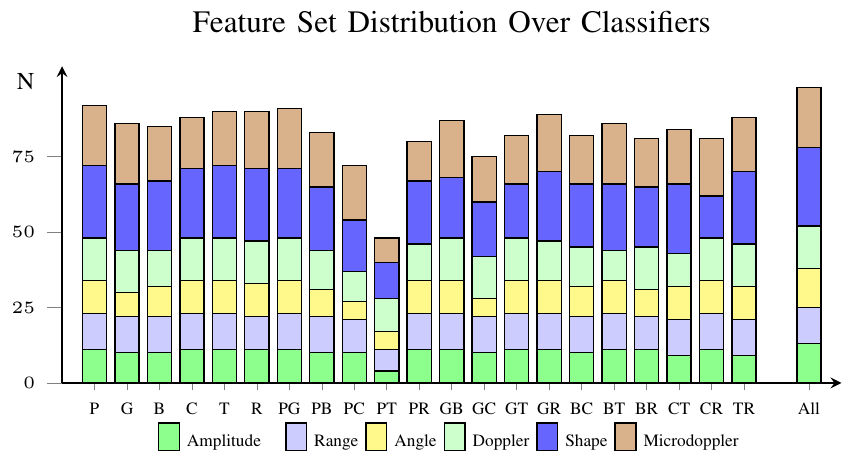}
\caption{Final feature distribution over classifiers in ensemble. OVA classifiers are identified by their corresponding ``one'' class, i.e., pedestrian (P), pedestrian group (G), bike (B), car (C), truck (T), or garbage (R). OVO classifiers are indicated likewise by two letters. The full feature categorization is displayed on the right for comparison.}
\label{fig:feat_dist}
\end{figure}

The distribution of categorized features over all classifiers in the ensemble is depicted in Fig. \ref{fig:feat_dist}.
A first observation of the displayed feature distribution is the expected tendency of classifiers to require a lower amount of features for simple classification problems.
It is clearly visible that, for instance, the \emph{pedestrian vs. truck} or \emph{pedestrian vs. car} classifier utilize a lot less features than the \emph{pedestrian vs. pedestrian group} or any one-vs-all classifier.
Despite this clear trend for facile subproblems, the variations between the number of utilized features in more difficult decision problems seem rather homogeneous.
The median number of utilized features per classifier is 85 which suggests, that many classifiers did remove a couple of features during backward elimination that had redundant information.
As there were no features that were always removed, no distinct conclusion can be made on this part.
However, this finding suggests that those classifiers might benefit from some additional features.

The influence of different feature categories on the individual classifiers can be summarized as follows:
The majority of classifiers uses all 12 range and all 14 Doppler features.
Even the \emph{pedestrian vs. truck} classifier with only 48 features in total still utilizes 11 Doppler features, despite massive removals all other categories.
Among the amplitude features, most classifiers settle at 10-11 from 13 features.
However, for \emph{angle}, \emph{shape}, and \emph{microdoppler} features a stronger correlation between total number of used features and dropped features from those categories can be observed.
This behavior is most distinct for \emph{angle} and \emph{shape} features.
A possible explanation for this, is that the angular information in the radar data is often not good enough.
As soon as the classifier has sufficiently enough good features, it starts to drop angle-based ones for better generalization performance.
This hypothesis is supported by the low angular resolution of radar sensors.
It is, therefore, expected that angular information will become more important when sensors with higher angular resolution are used.

\begin{figure}[tb]
\centering
\includegraphics{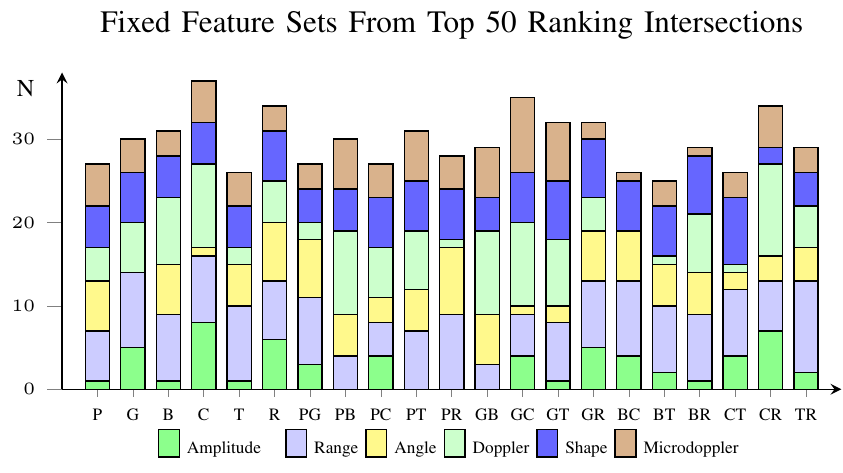}
\caption{Distributions of common features in the top 50 rankings of all classifiers in ensemble. OVA classifiers are identified by their corresponding ``one'' class, i.e., pedestrian (P), pedestrian group (G), bike (B), car (C), truck (T), or garbage (R). OVO classifiers are indicated likewise by two letters.}
\label{fig:fixed_feat}
\end{figure}

In order to also give an overview of the most important feature for each classifier, the merged set of the top 50 ranked features of both filter methods are depicted in Fig. \ref{fig:fixed_feat}.
Among these features, which are used as fixed set in the feature selection process, fewer clear trends are visible.
Each category has an average of $\approx5$ features except for \emph{amplitude} ($2.8$) and \emph{range} ($7.2$) features.
The lack of amplitude-based features in the fixed set suggests that a better representation of those values might be beneficial in the future.
Also, the constant high ratio of range-based and Doppler-based features in the final ensemble indicates that more features of those kinds could help the classification process even further.

\section{Conclusion}\label{sec:conclusion}
This articles examined the benefits of providing each classifier in a binarized multiclass classification network with its own unique feature set.
Motivated by the observation, that the high complexity of the combined one-vs-one and one-vs-all ensemble could not benefit from the design of a largely increased feature set (from 50 to 98), feature selection was performed individually for each single classifier in the ensemble.
Besides improving the classification performance, this technique also allows for getting a better understanding about the influence of individual features on specific classification subproblems.
In order to reduce the computational costs of these multiple feature selection procedures, a combined approach of a backward elimination routine guided by a heuristic to only evaluate one feature at a time was chosen.
The heuristic was made up by the combination of two feature ranking procedures, namely, \emph{Joint Mutual Information} and the Relief-based \emph{MultiSURF} algorithm.
The specialized classifier ensemble reaches an averaged F1 score of \SI{91.46}{\%} which is a strong performance boost when compared to previous methods.
Moreover, it was shown how the result aggregation stage of the ensemble can be manipulated to also detect hidden classes which have not been previously seen by the model during training.
Naturally, the overall classification performance of the model decreases when allowing to detect an additional class.
However, the ability to recognize objects from classes other than the ones seen in the training data, is a vital part towards autonomous driving.
For future work it is planned to enhance current results by applying high resolution signal processing techniques that allow to increase the radar's resolution in range, angle, and Doppler.
It is expected that classifier specific feature selection helps to find adequate compromises between an excellent classification performance and high computational demands of the signal processing techniques.

\section{Acknowledgment}
The research for this article has received funding from the European Union under the H2020 ECSEL Programme as part of the DENSE project, contract number 692449.

\bibliographystyle{IEEEtran}
\bibliography{IEEEabrv,mybibfile}

\newpage
\onecolumn
\section*{Appendix}
In Tab. \ref{tab:feats} all examined features are listed along with a short explanation.
The table also indicates the amount of features that are summarized by the corresponding row and the features categorization as used for the evaluation in Section \ref{sec:results} is indicated by abbreviations A (amplitude), R (range), P (Phi, i.e., angle), V (velocity), S (shape \& compactness), and D (spatial Doppler distribution).

\begin{table*}[htb!]
\renewcommand{\arraystretch}{1.3}
\caption{List Of Extracted Features.}
\label{tab:feats}
\centering
	\begin{tabular}{rllll}
		\hline
 		\textbf{\#} & \textbf{Name} & \textbf{Description} & \textbf{Amount} & \textbf{Group} \\
		\hline\hline \\ \hline
		\multicolumn{5}{c}{Features from all basic values (amplitude $A$, range $r$, angle $\phi$, ego-motion compensated radial velocity $v_r$)} \\
		\hline
		1 & Min & Minimum value in cluster & 4 & A/R/P/V \\
		2 & Max & Maximum value in cluster & 4 & A/R/P/V \\
		3 & Mean & Average value in cluster & 4 & A/R/P/V \\
		4 & MeanAbsDev & Mean absolute deviation (1st central absolute moment) & 4 & A/R/P/V \\
		5 & Var & Variance of value in cluster (2nd moment) & 4 & A/R/P/V \\
		6 & StdDev & Standard deviation of value in cluster & 4 & A/R/P/V \\
		7 & Skewness & 3rd statistical central moment & 4 & A/R/P/V \\
		8 & Kurtosis & 4th statistical central moment & 4 & A/R/P/V \\
		9 & Spread & Range between minimum and maximum value & 4 & A/R/P/V \\		
		\hline
		\multicolumn{5}{c}{Features calculated on specified groups} \\
		\hline
		10 & Log & Logarithmic value of \emph{meanAmplitude, rangeSpread, angleSpread, meanVelocity} & 4 & A/R/P/V \\
		11 & Sqrt & Square root of \emph{meanAmplitude, rangeSpread, angleSpread, meanVelocity} & 4 & A/R/P/V \\	
		12 & Quad & Squared value of \emph{meanAmplitude, rangeSpread, angleSpread, meanVelocity} & 4 & A/R/P/V \\
		13 & covEV & Eigenvalues of covariance matrix of $x$/$y$ and $x$/$y$/$v_r$/$A$ distributions & 6 & 2S/4D \\
		14 & covEV2 & Squared eigenvalues of \emph{covEV} & 6 & 2S/4D \\
		15 & con95axis & Axis lengths of 95\% confidence ellipses based on \emph{covEV} & 6 & 2S/4D \\
		\hline
		\multicolumn{5}{c}{Other features calculated from specific data} \\
		\hline
		16 & AmpSum & Sum of all amplitude values in cluster & 1 & A \\	
		17 & PhiSpreadComp & Angular cluster spread weighted by mean distance & 1 & P \\	
		18 & StdDevDoppler & Standard deviation of uncompensated Doppler velocities & 1 & V \\
		19 & fracStationary & Percentage of stationary detections in cluster & 1 & V \\
		20 & nDetects & Total amount of detections in cluster & 1 & S \\
		21 & nDetectsComp & Detection amount weighted by mean distance & 1 & S \\
		22 & nDetectsVolcan & Detection amount multiplied by volcanormal weighting function \cite{volca2017} & 1 & S \\
		23 & CorePoints & Ratio of core points to total amount of detections & 1 & S \\
		24 & MeanDist & Mean spatial distance of all detections in cluster & 1 & S \\ 	
		25 & clusterWidth & Maximum distance of any two detections in cluster & 1 & S \\
		26 & maxDistDev & Average target distance to the line corresponding to \emph{clusterWidth} & 1 & S \\
		27 & CBO & Cumulative binary occupancy of sectors on three concentric circles & 3 & 3S \\
		28 & RectHull & Area, perimeter and detection density of minimum rectangular hull & 3 & 3S \\
		29 & ConvexHull & Area, perimeter and detection density of convex hull & 3 & 3S \\
		30 & CircleFit & Radius of the best fitting circle & 1 & S \\
		31 & Circularity & Isoperimetric quotient of the convex hull & 1 & S \\
		32 & Compactness & Mean detection distance to cluster center & 1 & S \\
		33 & xyLinearity & Linear correlation between $x$ and $y$ values in cluster & 1 & S \\
		34 & rVrLinearity & Linear correlation between $r$ and $v_r$ values in cluster & 1 & D \\
		35 & phiVrLinearity & Linear correlation between $\phi$ and $v_r$ values in cluster & 1 & D \\
		36 & majorVrLinearity & Linear correlation between detection dilatation in major direction of $x$/$y$ conf. ellipse and $v_r$ & 1 & D \\
		37 & minorVrLinearity & Linear correlation between detection dilatation in minor direction of $x$/$y$ conf. ellipse and $v_r$ & 1 & D \\
		38 & rVrSpread & Ratio of range spread to $v_r$ spread & 1 & D \\
		39 & phiVrSpread & Ratio of angular spread to $v_r$ spread & 1 & D \\
		40 & majorVrSpread & Ratio of spatial spread along major axis of $x$/$y$ confidence ellipse to $v_r$ spread & 1 & D \\		
		41 & minorVrSpread & Ratio of spatial spread along minor axis of $x$/$y$ confidence ellipse to $v_r$ spread & 1 & D \\	
		\hline
		\multicolumn{3}{r}{\textbf{Total:}} & \textbf{98}\\
		\hline
\end{tabular}
\end{table*}

\end{document}